\documentclass[12pt, final]{l4dc2023}

\title[Transition Occupancy Matching]{Learning Policy-Aware Models for Model-Based Reinforcement Learning via Transition Occupancy Matching}
\usepackage{times}
\usepackage{algorithm}
\usepackage[noend]{algorithmic}
\usepackage{mathtools}
\usepackage{hyperref}

\newcommand{\para}[1]{\textbf{#1}}

\newcommand{\KL}{\mathrm{KL}}
\newcommand{\BE}{\mathbb{E}}
\newcommand{\mc}{\mathcal}

\newcommand{\jasonedit}[1]{\textcolor{black}{#1}}

\newcommand{\kedit}[1]{\textcolor{black}{#1}}

\usepackage[colorinlistoftodos]{todonotes}
\newcommand{\jdtext}[1]{\textcolor{black}{#1}}

\author{%
 \Name{Yecheng Jason Ma*} \Email{jasonyma@seas.upenn.edu}\\
 \Name{Kausik Sivakumar*} \Email{kausik@seas.upenn.edu}\\
 \Name{Jason Yan} \Email{jasyan@seas.upenn.edu}\\
  \Name{Osbert Bastani} \Email{obastani@seas.upenn.edu}\\
   \Name{Dinesh Jayaraman} \Email{dineshj@seas.upenn.edu}\\
 \addr University of Pennsylvania
}

\begin{document}

\maketitle

\begin{abstract}
Standard model-based reinforcement learning (MBRL) approaches fit a transition model of the environment to all past experience, but this wastes model capacity on data that is irrelevant for policy improvement. We instead propose a new ``transition occupancy matching'' (TOM) objective for MBRL model learning: a model is good to the extent that the current policy experiences the same distribution of transitions inside the model as in the real environment. We derive TOM directly from a novel lower bound on the standard reinforcement learning objective. To optimize TOM, we show how to reduce it to a form of importance weighted maximum-likelihood estimation, where the automatically computed importance weights identify policy-relevant past experiences from a replay buffer, enabling stable optimization. TOM thus offers a plug-and-play model learning sub-routine that is compatible with any backbone MBRL algorithm. On various Mujoco continuous robotic control tasks, we show that TOM successfully focuses model learning on policy-relevant experience and drives policies faster to higher task rewards than alternative model learning approaches. Code can be found on our project website: \href{https://penn-pal-lab.github.io/TOM/}{penn-pal-lab.github.io/TOM/}

\end{abstract}

\section{Introduction}
Model-based reinforcement learning (MBRL)~\citep{sutton1991dyna} is an effective paradigm for sample-efficient policy learning. In MBRL, an agent learns a dynamics model of its environment from its own experience. This learned dynamics model acts as a simulator, generating fictitious experience for policy optimization. The improved policy then generates new environment experiences, which are used to improve the dynamics model, and the cycle continues. 
MBRL's sample efficiency, coupled with breakthroughs in deep neural networks, have enabled impressive applications such as mastering Atari games and simulated robot control from pixels~\citep{hafner2019dream, hafner2019learning, hafner2020mastering, kaiser2019model}, in-hand dexterous manipulation~\citep{nagabandi2020deep}, and real-world robotics control~\citep{finn2017deep, ebert2018visual, wu2022daydreamer}.

The de facto standard approach to model learning trains the parameters of a neural network dynamics model by maximizing the likelihood of \textit{all} observed environment transitions (MLE). To see that this is inefficient,
consider a car on a road passing through rocky terrain. For the task of driving well, the agent need not know the complex dynamics of driving over the rocks; a model of the simple dynamics of the road surface would be sufficient. 
However, the MLE approach aims instead to learn a more comprehensive model, emphasizing all past experience equally, even when most of it is irrelevant to improving the current policy. For example, reinforcement learning (RL) policies acting largely randomly in early stages of training would mostly generate experiences of driving over rocks, and MLE models would continue to fit this data even after the policy has learnt to stay on the road.
This wastefulness is largely due to the \textit{objective mismatch}~\citep{lambert2020objective} between MLE model learning and optimal policy learning.  

To enable efficient training of task-relevant  models and  accelerate policy learning, we propose ``transition occupancy matching'' \textbf{(TOM)}. At a high level, TOM changes the model learning objective to focus more on environment transitions that the current policy can experience, generating ``policy-aware'' models well-suited for policy improvement. 
For example, in the car setting above, as the policy starts to spend more time on the road, the dynamics model adapts by fitting the ``footprint'' of this policy which now includes more on-road experience, driving faster policy improvement.

While the above procedure is intuitive, we derive TOM from first principles. Our \textbf{first} contribution is a novel lower bound to the standard RL objective, containing two parts:
(1) standard policy search through reward maximization with respect to a learned model, as in many prior MBRL approaches, and (2) a novel $f$-divergence model learning objective that aims to match the footprints of the current policy in the real environment and in the learned model.
\textbf{Second,} we overcome optimization challenges associated with the TOM model learning objective 
by showing that it is mathematically analogous to the offline imitation learning objective. With this insight, we adapt a recently proposed offline imitation approach~\citep{ma2022smodice} to reduce the TOM model-learning objective to \textit{importance-weighted} maximum likelihood model learning, where higher weights are assigned to past transitions that lie in the footprint of the current policy. 
This permits a stable, easy-to-implement optimization procedure. 
\textbf{Third, } we demonstrate that, when plugged into a standard MBRL system, TOM's model learning procedure induces better policies for various simulated robotics tasks faster than alternative approaches, by continuously focusing the model on the most relevant past experiences.

\begin{figure*}[t!]
\includegraphics[width=\linewidth]{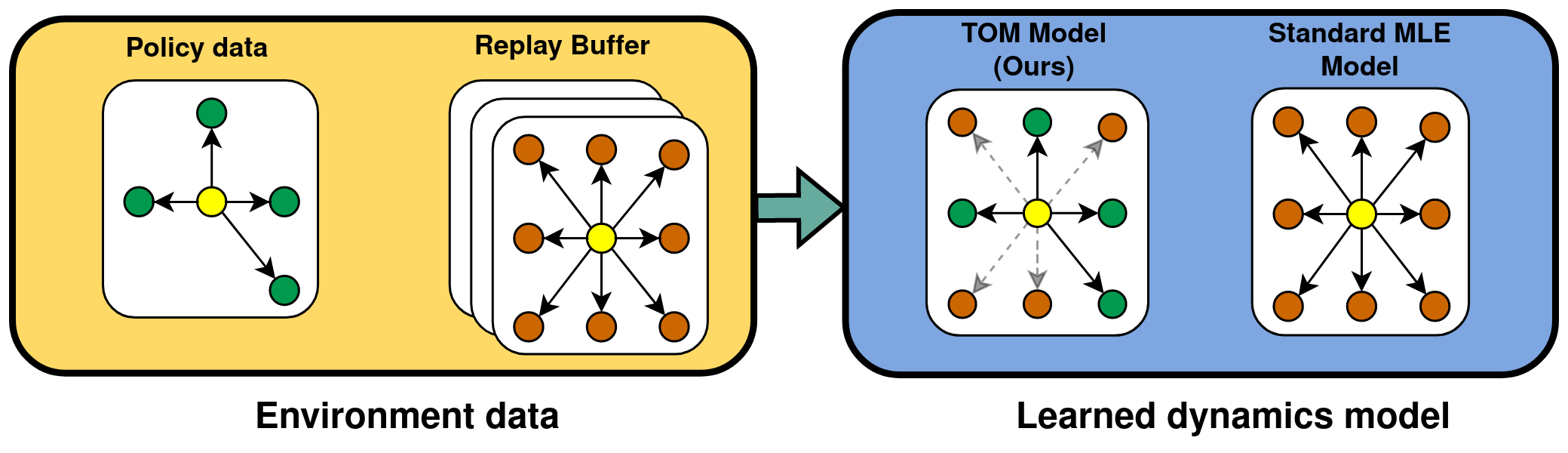}
\caption{Transition Occupancy Matching (TOM) enables learning a dynamics model that fits the policy's visitation distribution accurately to enable rapid policy improvement.}
\vspace{-0.5cm}
\label{figure:tom-concept-figure}
\end{figure*}

\section{Related Work}

Our work is broadly related to the objective mismatch problem~\citep{lambert2020objective} in MBRL. One prominent approach to address the objective mismatch problem is to make dynamics learning \textit{value-aware}~\citep{farahmand2017value, farahmand2018iterative, grimm2020value, farquhar2021self, voelcker2022value}. In particular, this line of work attempts to learn dynamics models that capture aspects of environment dynamics that impact accurate estimation of the value functions. \jasonedit{However, this paradigm entangles the policy's footprint with the task it is trying to solve, and often require well-shaped dense reward in the environment so that the value functions are not degenerate in the early iterations of policy optimization.}

Instead of focusing on the value function, our approach is more direct and \textit{policy-aware}~\citep{eysenbach2021mismatched, wang2022litm}, cognizant of the current policy's footprint without entangling it with the task it is solving. The closest work to ours is PMAC~\citep{wang2022litm}, which proposes to up-weight the most recent transitions in the replay buffer according to a hand-crafted weight schedule in regressing the dynamics model; however, this approach is heuristic in nature, and sensitive to hyperparameters. Furthermore, PMAC suffers from \textit{recency bias}; due to the inherent variance in policy optimization, the most recent transitions may be of low quality and not most relevant to improving the current policy, but PMAC would assign them high weights regardless. In contrast, TOM first establishes a lower bound to the true policy return objective in the transition occupancy space, then leverages techniques in dual reinforcement learning to derive a principled and optimal approach for assigning transition weights that is empirically effective.

\section{Background: Model-Based RL, State-Action Occupancies, and Bellman Flows} 

In this section, we will first go over the preliminaries for model-based reinforcement learning and then discuss the concept of state-action occupancy.

\vspace{0.03in}
\noindent \para{Model-based reinforcement learning.}
We consider an infinite horizon discounted Markov decision process (MDP)~\citep{puterman2014markov} $\mc{M}=(S,A,R,T, \mu_0, \gamma)$ where $S$ denotes its state space, $A$ its action space, $R$ the reward, $T(s,a)$ the transition function, $\mu_0(s)$ its initial state distribution, and $\gamma \in (0, 1]$ the discount factor. A policy $\pi:S \rightarrow \Delta(A)$ is a state-conditioned action distribution. The objective of RL is to find the policy $\pi$ that maximizes the discounted
return: 
\begin{equation}
\label{eq:rl-objective}
    J(\pi) \coloneqq \mathbb{E}_{s_0\sim \mu_0, a_t \sim \pi(\cdot \mid s_t), s_{t+1}\sim T(\cdot \mid s_t, a_t)}\left[\sum_{t=0}^{\infty} \gamma^t R(s_t.a_t) \right].
\end{equation}
We consider the online reinforcement learning setting, in which the agent directly interacts with the environment, collects new transitions $(s,a,r,s')$ and stores them in its replay buffer $D$. The agent's policy is updated using samples from $D$. \jasonedit{We define the replay buffer empirical policy as $\pi_D(a\mid s) := \frac{n(s,a)}{n(s)}$, where $n(\cdot)$ is the raw count of a state(-action) in $D$. }

Since the true dynamics $T$ is not known, MBRL builds an approximate dynamics model $\hat{T}$ which is learned from data. That is, a function approximator is built by designing $\hat{T}(s,a)$ as a probability distribution and by maximizing the likelihood of observing next state $s'$ given current state-action pair $(s,a)$ over transitions present in the collected replay buffer $(s,a,s') \sim D$. This can also be presented as minimizing the reverse KL divergence between transitions conditioned on samples in the replay buffer:

\begin{equation}
    \label{eq:kl-divergence}
    \jasonedit{\arg \min_{\hat{T}} \mathbb{E}_{D(s,a,s')}\KL\left(T(\cdot \mid s,a) \| \hat{T}(\cdot \mid s,a)\right) = \arg \min_{\hat{T}}\mathbb{E}_{D(s,a,s')} \left[\log \hat{T}(\cdot \mid s,a)\right]}.
\end{equation}

\noindent \para{State-Action Occupancy.}
The state-action occupancies (also known as stationary distribution) $d_T^\pi(s,a): \mc{S} \times \mc{A} \rightarrow [0,1]$ of policy $\pi$ is
\begin{equation}
\label{eq:pi-occupancies}
d_T^\pi(s,a) \coloneqq \;(1-\gamma) \sum_{t=0}^{\infty} \gamma^t \text{Pr}(s_t=s, a_t=a \mid s_0 \sim \mu_0, a_t \sim \pi(s_t), s_{t+1} \sim T(s_t,a_t)).
\end{equation}
This captures the relative state-action visitation frequencies of policy $\pi$ under dynamics $T$.\footnote{\jasonedit{Unless otherwise specified, we assume $d^\pi(s,a)$ is computed under the true dynamics $T$.}} The policy's state occupancies can be obtained by marginalizing over actions: $d^\pi(s) = \sum_a d^\pi(s,a)$. With this definition, we notice the following relationship between the policy and its occupancy distributions: 
$    \pi(a \mid s) = \frac{d^\pi(s,a)}{d^\pi(s)}.$

By construction, every policy's visitation distribution $d^\pi(s,a)$, must satisfy the single step transpose Bellman equation:
\begin{equation}
    \label{eq:bellman-flow-pi} 
    d^\pi(s,a) = (1-\gamma) \mu_0(s)\pi(a\mid s) + \gamma\pi(a\mid s) \sum_{\tilde{s}, \tilde{a}}T(s\mid \tilde{s}, \tilde{a})d(\tilde{s}, \tilde{a}).
\end{equation}
\jasonedit{This is known as the Bellman \textit{flow} constraint, which intuitively} restricts the ``flow'' of a policy's state-action distribution where each $d^\pi(s,a)$ must be expressed as a weighted sum. 
Conversely, a state-action occupancy distribution $d(s,a)$ needs to satisfy the Bellman flow constraint in order for it to be a valid $d^\pi(s,a)$ for some policy $\pi$:
\begin{equation}
\label{eq:bellman-flow}
\sum_a d(s,a) = (1-\gamma) \mu_0(s)+ \gamma  \sum_{\tilde{s}, \tilde{a}}T(s\mid \tilde{s}, \tilde{a})d(\tilde{s}, \tilde{a}), \forall s \in S, a\in A.
\end{equation}
Given $d^\pi$, one can express the RL objective~\eqref{eq:rl-objective} as:
\begin{equation}
\label{eq:rl-objective2}
J(\pi) = \frac{1}{1-\gamma} \mathbb{E}_{(s,a) \sim d^\pi(s,a)}[R(s,a)].
\end{equation} 
\jasonedit{Finally, we can reframe~\eqref{eq:rl-objective2} as a constrained optimization problem directly in the space of state-action occupancy distributions $d$ by incorporating the Bellman flow constraint~\eqref{eq:bellman-flow}:}
\begin{equation}
    \label{eq:primal-occupancy-problem}
    \begin{split}
    &\max_d \frac{1}{1-\gamma} \mathbb{E}_{(s,a) \sim d(s,a)}[R(s,a)] \\ 
    \mathrm{s.t.} \quad & \sum_a d(s,a) = (1-\gamma) \mu_0(s) + \gamma \sum_{\tilde{s}, \tilde{a}}T(s\mid \tilde{s}, \tilde{a})d(\tilde{s}, \tilde{a}), \forall s \in S, a\in A 
    \end{split}
\end{equation}

\section{Transition Occupancy Matching} 

We now develop our approach, transition occupancy matching (TOM). In Section~\ref{sec:tod}, we extend state-action occupancy to a new concept: transition occupancy, which formalizes the intuitive notion of the ``policy footprint'', motivated in the introduction. Next, in Section~\ref{sec:bound} we derive a novel lower bound to the RL objective that naturally suggests learning a policy-aware dynamics model as the key ingredient for model-based policy learning. Finally, Section~\ref{sec:algo} develops the full TOM algorithm by casting policy-aware model learning as an offline imitation problem.

\subsection{Extending State-Action Occupancy to ``Transition Occupancy''} \label{sec:tod}
Given a policy's state-action occupancy distribution $d^\pi_{\tilde{T}}(s,a)$ under a transition function $T$ \jasonedit{(with generality, $T$ in this section refers to any transition function and not necessarily the true environment dynamics)}, we define its \textit{transition occupancy distribution} (TOD) as:

\begin{equation}
\label{eq:transition-occupancy-distribution}
    d^\pi_T((s,a),s') := T(s'\mid s, a) d^\pi_T(s,a).
\end{equation}
Intuitively, $d^\pi_T((s,a),s')$ captures the relative frequency of any transition tuple $(s,a,s')$ that a policy visits under $T$. 
One immediate property of this definition is that we can back out the transition function as follows:
\begin{equation}
\label{eq:transition-from-occupancy}
    T(s' \mid s,a) := \frac{d^\pi_T((s,a),s')}{\sum_{s'} d^\pi_T((s,a),s')}, \forall \pi
\end{equation}

\jasonedit{Finally, we need to specify an analogous Bellman flow constraint for valid transition occupancies.} Note that the original Bellman flow constraint~\eqref{eq:bellman-flow} already contains a TOD term on the right: $T(s\mid \tilde{s}, \tilde{a})d^\pi(\tilde{s}, \tilde{a}) = d((\tilde{s},\tilde{a}),s')$. Therefore, by multiplying both sides of~\eqref{eq:bellman-flow} by $T(s'\mid s,a)$, we obtain the \textit{Bellman transition flow} constraint:
\begin{equation}
    \label{eq:bellman-transition-flow}
    \resizebox{\textwidth}{!}{$
    d^\pi_T((s,a),s') = (1-\gamma)\mu_0(s) T(s' \mid s,a) \pi(a\mid s) + \gamma     T(s' \mid s,a) \pi(a\mid s) \sum_{\tilde{s}, \tilde{a}} d^\pi_T((\tilde{s}, \tilde{a}), s), \forall(s,a,s')\in S\times A\times S $}
\end{equation}

\subsection{Policy-Aware Lower Bound via Transition Occupancy $f$-Divergence} \label{sec:bound}
With this new notion of ``transition occupancy distribution'' (TOD), we can rewrite the RL objective as: 
\begin{equation}
\label{eq:rl-objective3}
     J(\pi) = \jasonedit{\frac{1}{1-\gamma}}\BE_{d^\pi_T(s,a)}[R(s,a)] = \jasonedit{\frac{1}{1-\gamma}}\BE_{d^\pi_T((s,a), s')}[R(s,a)].
\end{equation}
Now, $\log J(\pi)$ permits a simple lower bound in terms of TODs:
\begin{equation}
\label{eq:lower-bound-derivation}
    \begin{split}
        \log J(\pi) = & \log \BE_{d^\pi_T((s,a), s')}[R(s,a)] + C\\
        = & \log \BE_{d^\pi_{\hat{T}}((s,a), s')}\left[\frac{d^\pi_T((s,a), s')}{d^\pi_{\hat{T}}((s,a), s')} R(s,a)\right] + C \\ 
        \geq & \BE_{d^\pi_{\hat{T}}((s,a), s')}\left[\log \frac{d^\pi_T((s,a), s')}{d^\pi_{\hat{T}}((s,a), s')}  + \log R(s,a)\right] + C \qquad \text{(by Jensen's inequality)} \\ 
        \geq & \underbrace{-\mathrm{D}_f (d^\pi_{\hat{T}}((s,a),s') \| d^\pi_T((s,a),s')}_{\jasonedit{\text{TOM model learning objective}}} + \underbrace{\BE_{d^\pi_{\hat{T}}((s,a), s')}[\log R(s,a) ]}_{\jasonedit{\text{standard RL within a learned model}}} + C,
    \end{split}
\end{equation}
\jasonedit{where $C$ is a constant that does not impact optimization and $f$ is any $f$-divergence that upper bounds the KL divergence.}

This lower bound \jasonedit{serves as a surrogate objective, and consequently (iteratively) maximizing it will lead to maximizing the true RL objective. More importantly, it consists of two expressions} that directly suggests a recipe for MBRL: (1) the first expression suggests trainining the dynamics model by minimizing the $f$-divergence between the distributions of real and fake policy rollouts, and (2) the second expression suggests optimizing the policy w.r.t. the learned model. \jasonedit{The second expression is just the RL objective~\eqref{eq:rl-objective3} with $\hat{T}$ instead of $T$, and can be optimized by any existing RL algorithm.} As such, TOM's technical contribution is a model learning sub-routine aimed at minimizing this $f$-divergence: 
\vspace{-0.3cm}
\begin{equation}
    \label{eq:tom-f}
    \min_{\hat{T}} \mathrm{D}_f (d^\pi_{\hat{T}}((s,a),s') \| d^\pi_T((s,a),s')
\end{equation}

\subsection{Optimizing the TOM objective}\label{sec:algo}
\jasonedit{We observe that optimizing~\eqref{eq:tom-f} requires estimating $d^\pi_{\hat{T}}((s,a),s')$; however, doing so requires rolling out $\pi$ in $\hat{T}$, which suffers from the compounding error of multi-step trajectory rollout in one-step dynamics models~\citep{lambert2022investigating} that makes the estimation inaccurate. Furthermore, this direct optimization approach fails to leverage the replay buffer $D$ that the agent has already collected. To circumvent these issues, we propose a practical algorithm that can learn a policy-aware model
tailored to the visitation distribution of $\pi$ without any additional samples from the real environment. The algorithm is derived by treating transition occupancy matching as an \textit{offline} imitation learning problem. We begin by illustrating the intuition of this reduction; then, we provide the technical derivation.}

\vspace{0.03in}
\noindent \para{Transition Occupancy Matching as Offline Imitation: An Analogy.~~} Below, we compare the TOM problem (left) and the well-known \textit{state-action} occupancy matching problem (right)~\citep{nachum2020reinforcement, ma2022smodice, ma2022vip, kim2022demodice}:
\begin{equation}
    \label{eq:tom-smodice-comparison}
    \min_{\hat{T}} \mathrm{D}_f (d^\pi_{\hat{T}}((s,a),s') \| d^\pi_T((s,a),s')
    \qquad \text{vs.} \qquad
    \min_\pi \mathrm{D}_f (d^\pi_{T}((s,a)) \| d^{\pi^*}_T((s,a))
\end{equation}
For state-action occupancy matching, the environment dynamics $T$ is fixed, and the goal is to learn a policy $\pi$ that matches the distribution of a target (optimal) policy $\pi^*$. And the TOM problem precisely \textit{reverses} the role of the policy $\pi$ and the dynamics model $T$. In this analogy, a ``transition function'' should map from ``state'' $(s,a)$, under ``action'' $s'$, to a distribution over new ``states'' $p(s', a' | s, a, s')=p(a'|s,a,s')$, which further reduces to $p(a'|s')$ under Markov assumptions. Note that this last distribution is precisely the policy $\pi$. In other words, the policy takes the place of the transition function and vice versa.

\vspace{0.03in}
\noindent \para{Regularized TOM For Offline Model Learning.~~} Given this analogy, we wish to derive a policy-aware model learning algorithm by adapting a suitable state-occupancy based imitation learning problem. In particular, we extend SMODICE~\citep{ma2022smodice} to allow TOM to learn the policy-aware dynamics model using just the replay buffer $D$, circumventing the issues laid out at the beginning of the section.
More specifically, we first regularize $d^\pi_{\hat{T}}((s,a),s')$ to the transition occupancy distribution $d^{\pi_D}_T((s,a),s')$ of the empirical policy $\pi_D$ via a choice of $f$-divergence, which \jasonedit{is a crucial step in} enabling learning $\hat{T}$ using solely the replay buffer $D$ without any additional simulated samples from $\hat{T}$ itself: \begin{equation}
\label{eq:tom-f-primal}
\begin{split}
    &\max_{d^\pi_{\hat{T}}} \quad  -\mathrm{D}_f(d^\pi_{\hat{T}}((s,a),s') \| d^\pi_T((s,a),s')) - \mathrm{D}_f(d^\pi_{\hat{T}}((s,a),s') \| d^{\pi_D}_T((s,a),s'))\\
    & \mathrm{s.t.} \quad \sum_{s'} d^\pi_{\hat{T}}((s,a),s') = (1-\gamma)\mu_0(s) \pi(a\mid s) + \gamma \pi(a\mid s) \sum_{\tilde{s}, \tilde{a}} d^\pi_{\hat{T}}((\tilde{s}, \tilde{a}), s), \forall (s,a) \in S \times A
\end{split}
\end{equation}
Here, we have incorporated the Bellman transition flow constraint~\eqref{eq:bellman-transition-flow}. %

\begin{algorithm}[t]%
\begin{small}
\caption{Transition Occupancy Matching}\label{alg:tom-deep-abbreviated}
\begin{algorithmic}[1]
\STATE \textbf{Require}: current policy $\pi$ and its environment rollout(s) $\tau$, replay buffer $D$
\STATE \textcolor{purple}{\texttt{// Discriminator Learning}}
\STATE Train discriminator $c^*(s,a,s')$ to separate policy-relevant transitions from others in the replay buffer \eqref{eq:adversarial-training} and derive $R(s,a,s')$. 
\STATE \textcolor{purple}{\texttt{// Q-Function Learning}}
\STATE Train policy-relevance Q-function $Q^*(s,a)$ using \eqref{eq:Q-problem}
\STATE \textcolor{purple}{\texttt{// Model Learning}}
\STATE Train policy-aware dynamics model $\hat{T}(s' \mid s, a)$ using~\eqref{eq:weighted_reg} %
\end{algorithmic}
\end{small}
\end{algorithm}

\vspace{0.03in}
\noindent \para{Offline Optimization In The Dual Form.~~}
Next, Eq~\eqref{eq:tom-f-primal} admits a simple dual problem that can be optimized using solely the replay buffer and whose optimal solution can be directly used to compute the primal optimal $d^\pi_{\hat{T}^*}$ and thereby $\hat{T}^*$: 
\begin{proposition}
\label{proposition:dual}
The dual problem to~\eqref{eq:tom-f-primal} is: 
\begin{equation}
\label{eq:Q-problem}
     \min_{Q \geq 0 } (1-\gamma) \mathbb{E}_{\mu_0, \pi}[Q(s,a)] + \mathbb{E}_{d^{\pi_D}_T(s,a,s')}\left[f_\star \left(\underbrace{\log \frac{d^\pi_T(s,a,s')}{d^{\pi_D}_T(s,a,s')}}_{:= r(s,a,s')} +\gamma \mathbb{E}_{\pi(a' \mid s')}[Q(s',a')] - Q(s,a)\right) \right]
\end{equation}
where $f_\star$ denotes the convex conjugate function of $f$. Given the optimal $Q^*$, $\forall(s,a,s')\in S\times A\times S$, the primal optimal solution satisfies
\begin{equation}
\label{eq:d-star}
d^\pi_{\hat{T}^*}((s,a),s') = d^{\pi_D}_T((s,a),s')f'_\star \left(r(s,a,s') + \gamma \mathbb{E}_{\pi(a' \mid s')}[Q^*(s',a')] - Q^*(s,a)\right)
\end{equation}
\end{proposition}
See Appendix~\ref{appendix:technical-derivations} for a proof. Conceptually, this dual problem learns a $Q$-function \kedit{(i.e., the dual variable)} that informs the relevance of state-action pairs for learning $d^\pi_{\hat{T}^*}((s,a),s')$ according to reward $r(s,a,s')$ (this is not the same as the task reward $R$), which is the log-ratio of the transition occupancy distributions of the current policy $\pi$ and the replay buffer empirical policy $\pi_D$; in Appendix~\ref{appendix:discriminator-training}, we detail how to estimate $r(s,a,s')$ in practice by training a transition discriminator. Crucially, this dual problem depends on only the replay buffer $D$, which includes samples from $\mu_0$, and the current policy $\pi$, without any requirement on further data collection. Therefore, in practice, we can approximate $Q^*$ by optimizing~\eqref{eq:Q-problem} using stochastic gradient descent (SGD) on $D$.

\jasonedit{Then, we can use \eqref{eq:d-star} to construct the optimal importance weights $\frac{d^\pi_{\hat{T}^*}((s,a),s')}{d^{\pi_D}_T((s,a),s')}$ and}
perform weighted regression~\citep{ma2022far}: 
\begin{equation}
\label{eq:weighted_reg}
\begin{split}
    \min_{\hat{T}} &- \mathbb{E}_{d^\pi_{\hat{T}^*}((s,a),s')}[\log \hat{T}(s' \mid s, a)] \\ 
    =& - \mathbb{E}_{d^{\pi_D}_T(s,a,s')}\left[ f'_\star \left(r(s,a,s') + \gamma \mathbb{E}_{\pi(a' \mid s')}[Q^*(s',a')] - Q^*(s,a)\right) \log \hat{T}(s' \mid s, a) \right] 
\end{split}
\end{equation}
Here, \jasonedit{the expectation depends only on the replay buffer distribution $\pi_D$, permitting supervised learning directly on $D$.} As such, we see that the dynamics model is still trained with MLE, but just on a different distribution that is \textit{policy-aware}; equivalently, the model is learned via \textit{behavior cloning} (BC) on the current policy's transition occupancy distribution. As such, TOM retains the training stability of supervised learning, while simultaneously ensuring policy-awareness that enables optimizing a well-defined lower bound to the true RL objective and improved sample efficiency. The TOM algorithm is summarized in Alg.~\ref{alg:tom-deep-abbreviated}, and a full version is in Alg.~\ref{alg:tom-practical}, Appendix~\ref{appendix:pseudocode}.

\section{Experiments}
To evaluate TOM, we measure policy rewards and sample efficiency of policy learning, and also investigate whether TOM successfully focuses the model on policy-relevant samples.

\subsection{Offline Linear MBRL on Road-And-Rocks} 
\begin{figure}[t]
    \centering
    \includegraphics[width=\linewidth]{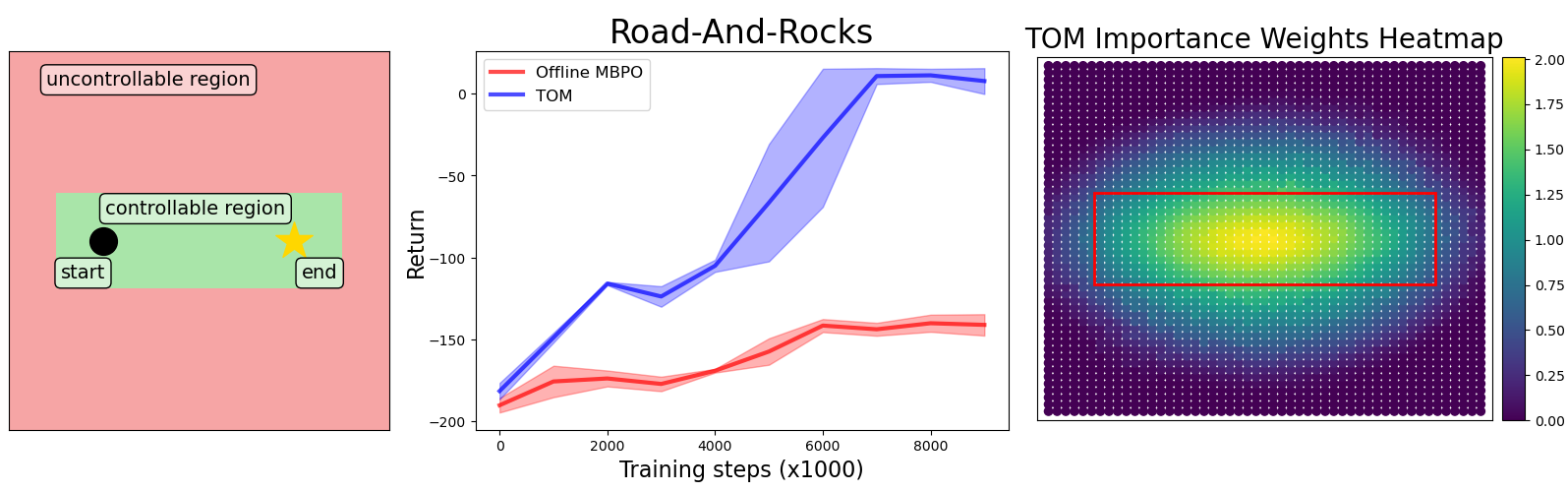}
    \caption{\jasonedit{Road-And-Rocks toy environment (left). TOM substantially outperforms (offline) MBPO (middle), and is doing so because it is able to correctly assign higher importance weights to the controllable region (right).}}
    \vspace{-0.5cm}
    \label{fig:TOM_rect_fig}
\end{figure}

We first evaluate TOM on a ``road-and-rocks'' task, based on the introduction example, shown in Figure~\ref{fig:TOM_rect_fig} (left). A ``car'' agent can steer in any direction. In the green ``road'' region, these actions uniformly produce the desired motions. In the red ``rocks'' region, the effects are different at each location, simulating the complex dynamics of driving on rocky terrain. The task is to drive on the road from a start to a goal position. For this toy experiment, we train a linear dynamics model. Off-road dynamics are highly non-linear, however the linear model will suffice if focused on the correct task-relevant on-road data.  To isolate the effect of model learning from the quality of data collected by online exploration~\citep{lambert2022investigating}, we consider an offline model-based reinforcement learning~\citep{yu2020mopo, kidambi2020morel} setting; policy optimization takes place using only pre-recorded ``offline'' environment transitions without any further data collection. The offline data includes random exploratory dataset over the entire map, \jasonedit{plus a small amount of ``expert'' trajectories
that successfully complete the task; this resembles practical use cases where the dataset provides high coverage of the state space but is mostly sub-optimal and task-irrelevant.} For TOM, one such expert trajectory is treated as the current policy footprint.  %

As expected, Figure~\ref{fig:TOM_rect_fig} (middle) shows vastly better environment reward curves when replacing MBPO model-learning with TOM. The inferred importance weights for transitions in various parts of the heatmap, shown in Figure~\ref{fig:TOM_rect_fig} (right), confirm that TOM successfully focuses the model on the task-relevant road regions, de-emphasizing the highly non-linear rocky regions.

\subsection{Online Deep MBRL on Simulated Robotics Tasks}
\jdtext{We now move from the toy road-and-rocks environment above to online deep MBRL evaluations in standard MuJoCo continuous control benchmarks~\citep{todorov2012mujoco}.}

\begin{figure}[t]
    \centering
    \includegraphics[width=\linewidth]{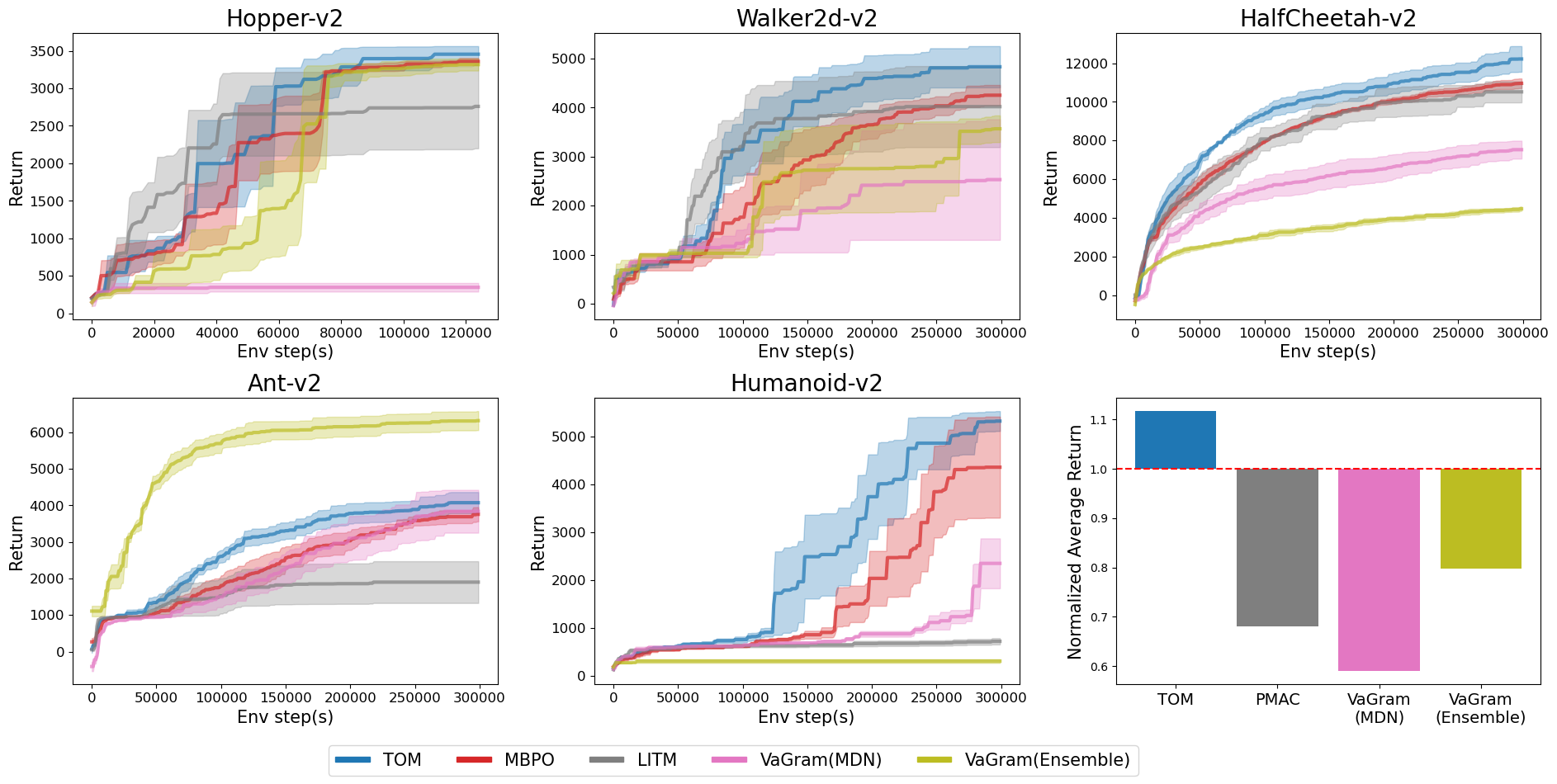}
    \caption{Return plots on Mujoco environments.
    }
    \vspace{-0.5cm}
    \label{fig:TOM_online_experiment}
\end{figure} 

\noindent \para{Baselines.} 
\jdtext{We compare TOM within the widely used MBPO framework~\citep{janner2019trust} to standard \textbf{MLE} model learning, and two representative recent approaches that target the MBRL objective mismatch problem. \textbf{PMAC}~\citep{wang2022litm} is a policy-aware approach that heuristically upweights recent transitions in the replay buffer following an exponentially decaying schedule. \textbf{VaGram}~\citep{voelcker2022value} is a value-aware approach, that reweights modeling errors along different state dimensions by the gradient of the value function. For all algorithms, we model the dynamics using mixture density networks architectures (MDN)~\citep{bishop1994mixture} that model the output distribution as a mixture of Gaussians.
For VaGram, which originally uses an ensemble of uni-modal Gaussian MLPs~\citep{chua2018deep}, we also compare to that version (``\textbf{VaGram (Ensemble)}'') and denote the MDN version of VaGram as ``\textbf{VaGram (MDN)}''.}

\vspace{0.03in}
\noindent \para{Environments and Training Details.} On five standard Mujoco environments: Hopper, Walker, HalfCheetah, Ant, and Humanoid, we train policies for up to 300k environment steps. 
We inherit standard MBPO hyperparameters for TOM and all our model-learning baselines. 
We evaluate all algorithms for 4 seeds and report the cumulative maximum average return over 10 test rollouts achieved during training. See Appendix~\ref{appendix:implementation-details} for all  implementation details.%

\vspace{0.03in}
\noindent \para{Reward Curves.~~}%
The reward curves in Figure \ref{fig:TOM_online_experiment} show that TOM performs the best in aggregate across \kedit{four out of five} environments, in terms of both sample efficiency and final policy performance. TOM's gains are larger in environments with more complex dynamics, such as Walker, HalfCheetah, and Humanoid, where judicious use of model capacity is more critical. 
Notably on Humanoid, TOM learns a policy that achieves at least 60\% more reward than plain MBPO. 
PMAC is able to initially outpace TOM on the simpler environments such as Hopper and Walker; however, this advantage is quickly erased as training continues, and PMAC converges to a worse return than TOM. Furthermore, on the more difficult environments such as Ant and Humanoid, PMAC is highly sub-optimal and reaches much lower asymptotic performance. These results suggest that a more principled approach TOM is needed in order to be correctly policy-aware.

VaGram, on the other hand, fluctuates in performance with large variance across the two choices of dynamics model and state dimensions. VaGram(Ensemble) struggles in all environments except Ant; notably, it cannot learn any meaningful behavior on the most difficult Humanoid; VaGram(MDN) generally converges to sub-optimal policies compared to TOM and fails at even the simplest Hopper task. We hypothesize that VaGram's usage of the first-order value gradient information for guiding the dynamics learning may be quite sensitive to the choice of dynamics model.

\begin{figure}[t]
    \centering
    \includegraphics[width=0.9\linewidth]{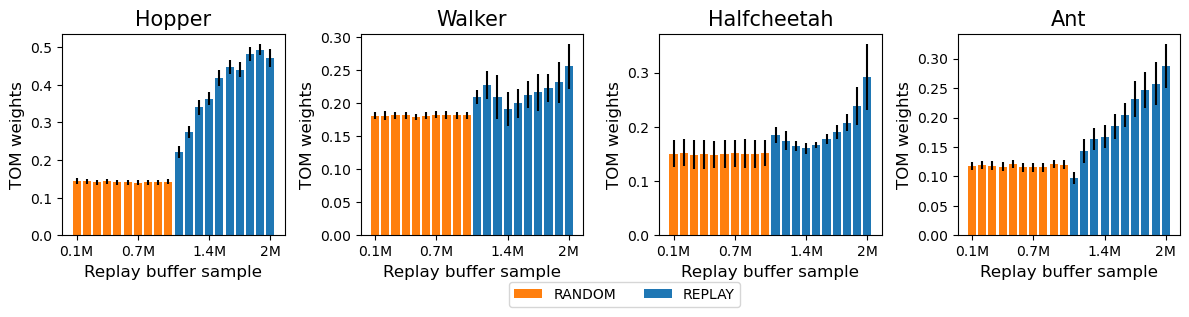}
    \vspace{-0.2cm}
    \caption{TOM transition importance weights.}
    \vspace{-0.5cm}
    \label{fig:tom_progression}
\end{figure}

\noindent \para{Analysis of TOM Importance Weights.}
It is clear above that TOM performs well relative to baselines in these tasks, but does it do so for the right reasons? In other words, does it actually focus the model on the right data?
To investigate this systematically, we manually curate a carefully ordered replay buffer, ordering it such that the first 50\% of the data contains random transitions, and the next 50\% contains data from various ordered stages in the training progress of a soft-actor critic (SAC)~\citep{haarnoja2018soft} policy agent, as provided by the D4RL dataset~\citep{fu2021d4rl}. We fix the fully trained ``expert'' SAC agent as the policy of interest, and run TOM model learning fully offline. We should expect good importance weights to be uniformly low over the first half of the replay buffer and then increase as we progress through the second half. And indeed, Figure~\ref{fig:tom_progression} clearly shows these trends for the average TOM weights in various chunks of the replay buffer.

It is also pertinent that TOM weights do not only rise at the very end of the replay buffer, when data is near-optimal; instead, they start to rise earlier, sometimes even non-monotonically, suggesting that the recency heuristic by \jasonedit{PMAC}, is not the sole determinant of relevance for model learning. In Appendix~\ref{appendix:online-progression}, we show that this also commonly occurs during the online MBRL experiments: TOM-assigned weights are quite frequently higher for older data than for newer ones.

\section{Conclusion} 
We have introduced Transition Occupancy Matching (TOM), a principled policy-aware model learning approach to address the objective mismatch challenge in model-based reinforcement learning. TOM introduces the notion of transition occupancy and derives a simple lower bound to the reinforcement learning objective, which permits casting learning a policy-aware dynamics model as learning the optimal importance weights for weighted regression model updates. The importance weights are derived from the theory of dual reinforcement learning, and TOM's practical implementation is modular and compatible with any MBRL algorithm that implements MLE regression-based model learning. On the standard suite of Mujoco tasks, TOM improves the learning speed of a standard MBRL algorithm while achieving significantly higher asymptotic performance compared to non-policy aware methods.

\section*{Acknowledgments}
This work was supported by an ONR award N00014-22-1-2677 to DJ.

\newpage 
\bibliography{l4dc_refs} 

\newpage 
\appendix

\section{Technical Derivations}
\label{appendix:technical-derivations}
In this section, we provide the omitted technical derivations in the main text.

\subsection{Proof of Proposition~\ref{proposition:dual}}
We begin by stating an assumption on the strict feasibility of the Bellman transition flow constraint in~\eqref{eq:tom-f-primal}.
\begin{assumption}
\label{assumption:strict-feasibility}
\rm
There exists at least one $d(s,a,s')$ such the Bellman transition flow constraint is satisfied and $\forall s \in \mathcal{S}, d(s) > 0$.
\end{assumption}
This assumption is mild and typically satisfied for any practical MDP in which every state is reachable from the initial state distribution. 

Now, we first write down the Lagrangian of~\eqref{eq:tom-f-primal}:
\begin{equation}
    \begin{split}
        \max_{d^\pi_{\hat{T}}} \min_{Q} & -\mathrm{D}_f(d^\pi_{\hat{T}}((s,a),s') \| d^\pi_T((s,a),s')) - \mathrm{D}_f(d^\pi_{\hat{T}}((s,a),s') \| d^{\pi_D}_T((s,a),s'))\\ 
        &+ \sum_{s,a} Q(s,a) \left((1-\gamma)\mu_0(s) \pi(a\mid s) + \gamma \pi(a\mid s) \sum_{\tilde{s}, \tilde{a}} d^\pi_{\hat{T}}((\tilde{s}, \tilde{a}), s) - \sum_{s'}d^\pi_{\hat{T}}((s,a),s') \right)
    \end{split}
\end{equation}
We use the following two identities to simplify the objective:
\begin{equation}
    \sum_{s,a}Q(s,a) \left(\sum_{s'}d^\pi_{\hat{T}}((s,a),s')\right) = \mathbb{E}_{d^\pi_T((s,a),s')}[Q(s,a)]
\end{equation}
and 
\begin{equation}
\sum_{s,a} Q(s,a) \left(\gamma \pi(a\mid s) \sum_{\tilde{s}, \tilde{a}} d^\pi_{\hat{T}}((\tilde{s}, \tilde{a}), s)\right) = \gamma \mathbb{E}_{d^\pi_{\hat{T}}((s,a),s')}\mathbb{E}_{\pi(a'\mid s')} [Q(s',a')],
\end{equation}
which follow from standard algebraic manipulations.
Using these identities and the strict feasibility assumption, strong duality enables switching the order of optimization and simplifies the objective to
\begin{equation}
\label{eq:tom-f-dual-intermediate}
\begin{split}
    \min_{Q } \max_{d^\pi_{\hat{T}}\geq 0} & (1-\gamma) \mathbb{E}_{\mu_0, \pi}[Q(s,a)] + \mathbb{E}_{d^\pi_{\hat{T}}(s,a,s')}\left[ \underbrace{\log \frac{d^\pi_T(s,a,s')}{d^{\pi_D}_T(s,a,s')}}_{:= R(s,a,s')} + \gamma \mathbb{E}_{\pi(a' \mid s')}[Q(s',a')] - Q(s,a) \right]\\ 
    &- \mathrm{D}_f(d^\pi_{\hat{T}}((s,a),s') \| d^{\pi_D}_T((s,a),s'))
\end{split}
\end{equation}

Then, we recognize that the inner maximization is precisely the Fenchel conjugate of

\begin{equation} \mathrm{D}_f(d^\pi_{\hat{T}}((s,a),s') \| d^{\pi_D}_T((s,a),s'))
\end{equation} 
at $R(s,a,s') + \gamma \mathbb{E}_{\pi(a' \mid s')}[Q^*(s',a')] - Q^*(s,a)$, which allows us to reduce~\eqref{eq:tom-f-dual-intermediate} to the Fenchel \textit{dual} problem of~\eqref{eq:tom-f-primal}:
\begin{equation}
\label{eq:tom-f-dual}
    \min_{Q } (1-\gamma) \mathbb{E}_{\mu_0, \pi}[Q(s,a)] + \mathbb{E}_{d^{\pi_D}_T(s,a,s')}\left[f_\star (R(s,a,s') +\gamma \mathbb{E}_{\pi(a' \mid s')}[Q(s',a')] - Q(s,a) \right]
\end{equation}
Then, leveraging Lemma 3 from~\citet{ma2022smodice}, it follows that
\begin{equation}
\label{eq:optimal-ratio} d^\pi_{\hat{T}^*}((s,a),s') = d^{\pi_D}_T((s,a),s') 
f'_\star \left(r(s,a,s') + \gamma \mathbb{E}_{\pi(a' \mid s')}[Q^*(s',a')] - Q^*(s,a)\right)
\end{equation}

\subsection{Discriminator Training}
\label{appendix:discriminator-training}
In practice, $r(s,a,s')$ can be estimated by training a discriminator  $c:\mc{S \times A \times S} \rightarrow (0,1)$ that distinguishes transitions from $\pi$ and $\pi_D$: 
\begin{equation}
    \label{eq:adversarial-training}
    \min_c \BE_{(s,a,s') \sim d^\pi_T(s,a,s')}\left[\log c(s,a,s') \right] + \BE_{d^{\pi_D}_T(s,a,s')}\left[\log 1-c(s,a,s') \right]
\end{equation}
The optimal discriminator is $c^\star(s,a,s') = \frac{d^\pi_T(s,a,s')}{d^\pi_T(s,a,s')+d^{\pi_D}_T(s,a,s')}$~\citep{goodfellow2014generative}, so we can use $r(s,a,s') = -\log \left(\frac{1}{c^\star(s,a,s')}-1\right)$.

\section{Pseudocode}
\label{appendix:pseudocode}
\begin{algorithm}[H]%
\caption{ Transition Occupancy Matching for continuous control}
\label{alg:tom-practical}
\begin{algorithmic}[1]
    \STATE Initialize policy $\pi_\phi$, predictive model $p_\omega$,discriminator $c_\psi$, Q function $Q_\theta$, environment dataset $\mathcal{D}_\text{env}$, model dataset $\mathcal{D}_\text{model}$, Current policy pool dataset $\mathcal{D}_\text{pol}$,choice of $f$-divergence $f$ 
    \FOR{$N$ epochs}
        \STATE \textcolor{purple}{\texttt{// Train Expert Discriminator}}
        \STATE Train Discriminator $c_\psi$ using $\mathcal{D}_\text{pol}$ and $\mathcal{D}_\text{env}$
        \STATE \textcolor{purple}{\texttt{// Train Lagrangian Q Function}}
        \FOR{\text{$U$ iterations}}
            \STATE Sample minibatch data from environment pool $\{s_t^i, a_t^i, s_{t+1}^i\}_{i=1}^N \sim \mathcal{D}_\text{env}$ and $\{s_0^i\}_{i=1}^M \sim \mathcal{D}_\text{env}(\mu_0)$
            \STATE Obtain reward: $R_i = c_\psi(s^i_t,a^i_t,s^i_{t+1}), i=1,...,N$
            \STATE Compute value objective \\$\mc{L}(\theta) \coloneqq (1-\gamma) \frac{1}{M}\sum_{i=1}^M V_\theta(s_0^i) + \frac{1}{N} f_\star\left(R_i + \gamma V(s_{t+1}^i)-Q(s^i_t,a^i_t)\right)$
            \\where $V(s_{t+1}) = \frac{1}{P}\Sigma_p Q(s_{t+1},a^p_{t+1})$ where $a^p_{t+1} \sim \pi_\phi(s_{t+1})$  
            \STATE Update $Q_\theta$ using SGD: $Q_\theta \leftarrow Q_\theta - \eta_Q \nabla \mc{L}(\theta)$
        \ENDFOR
        \STATE \textcolor{purple}{\texttt{// Model Learning}}
        \FOR{\text{$H$ iterations}}
            \STATE \textcolor{purple}{\texttt{// Compute Optimal Importance Weights for all env pool samples}}
            \STATE Compute $\xi^*(s_t^i,a_t^i,s^i_{t+1}) = f'_\star\left(R(s_t^i,a_t^i,s_{t+1}^i) + \gamma V(s_{t+1}^i) - Q(s_t^i,a_t^i) \right), i=1,...,N$
            \STATE Sample minibatch data from environment pool accoriding to the weights $\{s_t^i, a_t^i, s_{t+1}^i\}_{i=1}^N \sim \mathcal{D}_\text{env}$ according to $\xi^*(s_t^i,a_t^i,s^i_{t+1})$
            \STATE Obtain reward: $R = c_\psi(s^i_t,a^i_t,s^i_{t+1}), i=1,...,N$
            
            \STATE \textcolor{purple}{\texttt{// Weighted Regression}}
            \STATE Train dynamics model $p_\omega$ on sampled minibatch %
        \ENDFOR
        
        \FOR{$E$ steps}
            \STATE Take action in environment according to $\pi_\phi$; add to $\mathcal{D_\text{env}}$
            \FOR{$M$ model rollouts}
                \STATE Sample $s_t$ according to weights $\xi^*(s_t^i,a_t^i,s^i_{t+1})$ from $\mathcal{D}_\text{env}$
                \STATE Perform $k$-step model rollout starting from $s_t$ using policy $\pi_\phi$; add to $\mathcal{D_\text{model}}$
            \ENDFOR
            \FOR{$G$ gradient updates}
                \STATE Update policy parameters on model data: $\phi \leftarrow \phi - \lambda_\pi \hat{\nabla}_\phi J_\pi(\phi, \mathcal{D}_\text{model})$
            \ENDFOR
        \ENDFOR
        \STATE Update $\mathcal{D}_\text{pol}$ for the current trained policy
    \ENDFOR
\end{algorithmic}
\label{alg:TOM_pscode}
\end{algorithm}

\section{Implementation Details}
\label{appendix:implementation-details}

\subsection{Hyperparameters and Architecture}
\label{section:hyperparameters}
We standardize hyperparameters across all experiments and environments; they are listed in Table~\ref{tab:hyperparameters}.

In terms of network architectures, the dynamics value function $Q_\theta$ is implemented as a simple 2-layered ReLU network each with 256 neurons in the hidden dimension. The discriminator $c_\psi$ has the same architecture as the value function but with Tanh as its activation. 
The dynamics model $p_\omega$ is a sigmoid activated 4 layer neural network with 200 hidden neurons in each layer. In Humanoid, we use 400 hidden neurons in each layer for the dynamics model.

We employ Soft-Actor-Critic (SAC) ~\citep{haarnoja2018soft} for policy optimization and use default architecture in a publicly released implementation. 

\begin{center}
\begin{tabular}{c|c }
  \hline
  Hyperparameter & Value\\
  \hline
  Optimizer &   Adam ~\cite{kingma2014adam}\\
  Learning Rate &   3e-4\\
  Divergence    &   $\chi^2$-divergence\\
  Discriminator update steps per iteration & 100\\
  Discriminator batch size  &   256\\
  Value network update steps per iteration & 1000\\
  Value network batch size  & 256\\
  Dynamics model update steps per iteration & 30\\
  Dynamics model batch size     &   256\\
  Policy network batch size     &   256\\
  Current policy buffer capacity  &   1000\\
  Replay buffer capacity             &   1000000\\
  Rollout batch size            &   100000\\
  Model rollout step(s)         &   1\\
  PMAC decay rate               &   0.996\\    
  \hline
\end{tabular}
\label{tab:hyperparameters}
\end{center}

\subsection{PMAC}
\label{section:pmac_implementation}
In this section we describe the PMAC \citep{wang2022litm} baseline implementation details. Like \citep{wang2022litm}, we employ a sampling based regression approach, where the importance weights for the replay buffer are interpreted as the probability mass for each sample. These importance weights are assigned via an exponential up-weighting heuristic that pays more attention to recently collected transitions. Specifically, in this implementation the transitions collected by historical data are decayed by 0.996 while the transitions collected by the current policy rollouts share weight of 0.004. We implement PMAC on the same base MBPO implementation that TOM is built on, providing fair comparison of the different model learning approaches.

\subsection{Current policy buffer}
\label{section:current_policy_buffer}
We approximate the current policy buffer with the last 1000 transitions collected by the agent in the environment. This circumvents the need of collecting extra data by employing the current policy.

\section{Online progression}
\label{appendix:online-progression}
To demonstrate that TOM does not only pay attention to recent transitions, we capture the importance weights calculated by TOM and plot them with respect to the sequential order of collected transitions. The y-axis represents the TOM importance weights and the x-axis is sequential order of replay buffer transitions averaged in 100 bins
\begin{figure}[H]
    \centering
    \includegraphics[width=\linewidth]{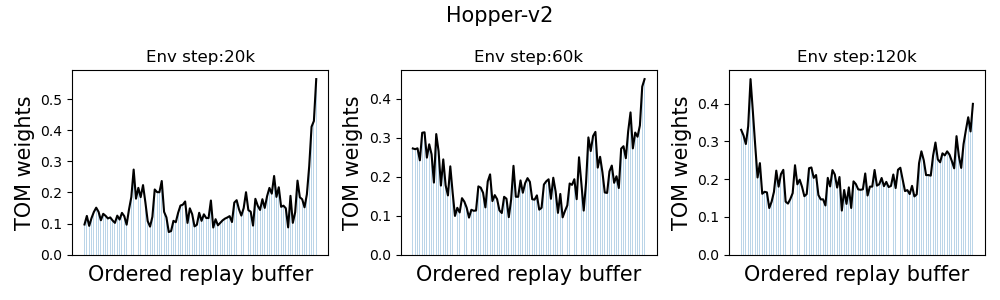}
    \caption{Hopper online buffer weights - total steps (125k)}
    \label{fig:hop_online}
\end{figure}
\vspace{0.00mm} 
\begin{figure}[H]
    \centering
    \includegraphics[width=\linewidth]{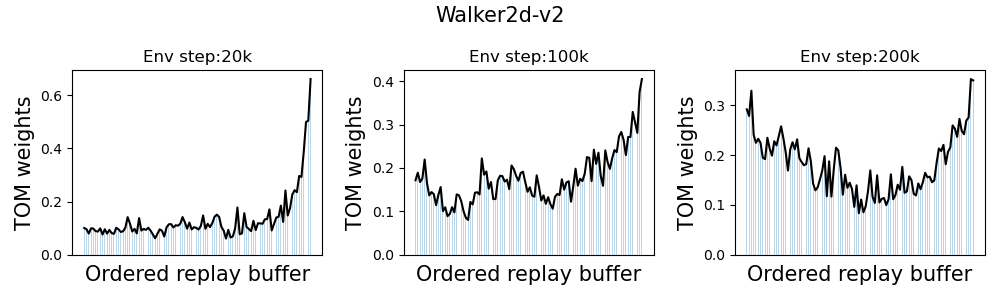}
    \caption{Walker online buffer weights- total steps (300k)}
    \label{fig:walk_online}    
\end{figure}
\vspace{0.00mm} 
\begin{figure}[H]
    \centering
    \includegraphics[width=\linewidth]{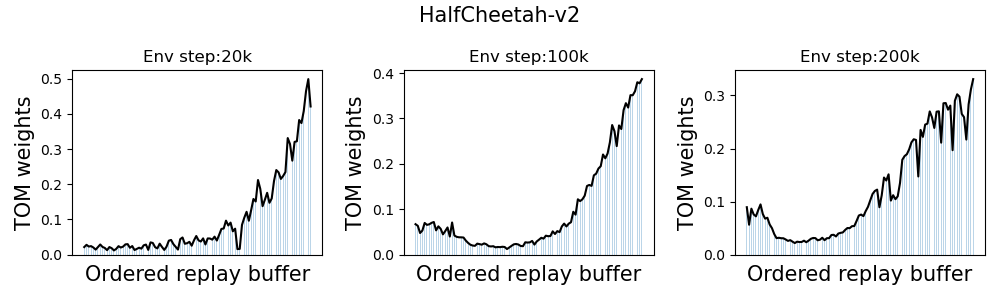}
    \caption{HalfCheetah online buffer weights- total steps (300k)}
    \label{fig:hc_online}
\end{figure}
\vspace{0.00mm} 
\begin{figure}[H]
    \centering
    \includegraphics[width=\linewidth]{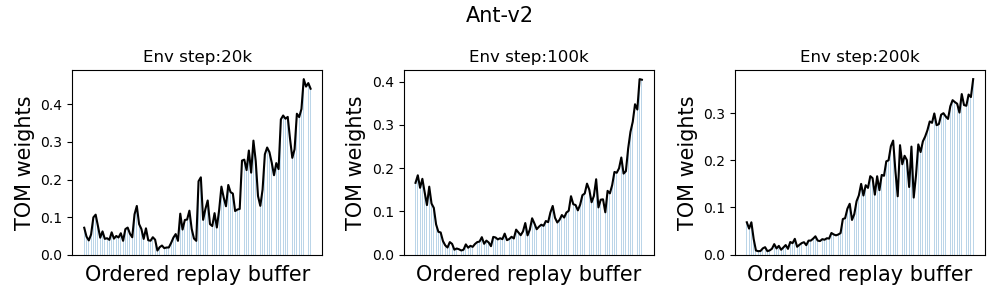}
    \caption{Ant online buffer weights- total steps (300k)}
    \label{fig:ant_online}
\end{figure}
\vspace{0.00mm} 
\begin{figure}[H]
    \centering
    \includegraphics[width=\linewidth]{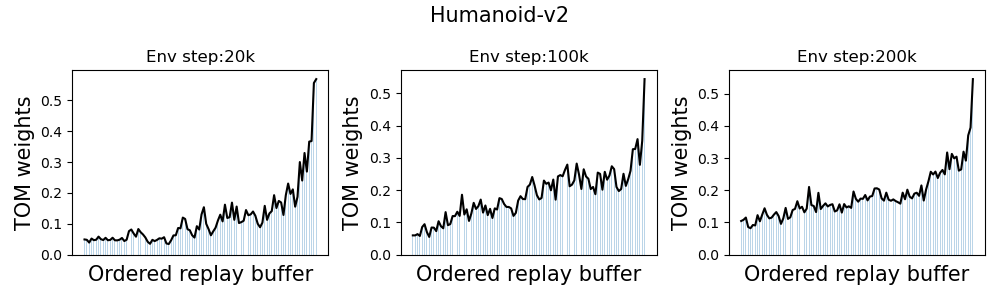}
    \caption{Humanoid online buffer weights- total steps (300k)}
    \label{fig:hum_online}
\end{figure}
These results show that TOM's importance weights are not high only at the very end of the replay buffer where data is near-optimal. It often pays attention to transitions that were collected at earlier stages. This trend is clearly noticeable in Figure \ref{fig:hop_online} (step 120k) and Figure \ref{fig:walk_online} (step 200k), where TOM pays more pays more attention to earlier transitions in the replay buffer, clearly suggesting that paying attention to recent transitions is not the only relevant factor for learning a policy-aware dynamics model.

\end{document}